\documentclass[letterpaper, 10 pt, conference]{ieeeconf}  

\usepackage{graphics} 
\usepackage{epsfig} 
\usepackage{mathptmx} 
\usepackage{times} 
\usepackage{amsmath} 
\usepackage{amssymb}  

\usepackage{subfig}
\usepackage{multirow}
\usepackage{tikz}
\usepackage{comment}
\usepackage{color}
\usepackage{footnote}
\usepackage{url}
\usepackage{fancyhdr}
\usepackage[left=1.75cm,right=1.75cm,top=1cm,bottom=2.5cm]{geometry}

\title{\Huge FreDSNet: Joint Monocular Depth and Semantic Segmentation with Fast Fourier Convolutions} 

\date{September, 2022}

\author{Bruno Berenguel-Baeta and Jesús Bermúdez-Cameo and Jose J. Guerrero}

\renewcommand{\headrulewidth}{0pt}
\renewcommand{\footrulewidth}{1pt}
\begin{document}

\maketitle
\thispagestyle{fancy}

\let\thefootnote\relax\footnote{All authors are with Instituto de Investigacion en Ingenieria de Aragon, University of Zaragoza, Spain}
\let\thefootnote\relax\footnote{Corresponding author: {\tt\small berenguel@unizar.es}}
\let\thefootnote\relax\footnote{A final version of this article can be found at \url{https://doi.org/10.1109/ICRA48891.2023.10161142}}

\begin{abstract}
In this work we present FreDSNet, a deep learning solution which obtains semantic 3D understanding of indoor environments from single panoramas.
Omnidirectional images reveal task-specific advantages when addressing scene understanding problems due to the 360-degree contextual information about the entire environment they provide. However, the inherent characteristics of the omnidirectional images add additional problems to obtain an accurate detection and segmentation of objects or a good depth estimation. To overcome these problems, we exploit convolutions in the frequential domain obtaining a wider receptive field in each convolutional layer. These convolutions allow to leverage the whole context information from omnidirectional images. FreDSNet is the first network that jointly provides monocular depth estimation and semantic segmentation from a single panoramic image exploiting fast Fourier convolutions. Our experiments show that FreDSNet has similar performance as specific state of the art methods for semantic segmentation and depth estimation.
FreDSNet code is publicly available in \url{https://github.com/Sbrunoberenguel/FreDSNet}
\end{abstract}

\section{Introduction}

Understanding 3D indoor environments is a hot topic in computer vision and robotics research \cite{naseer2018indoor}\cite{zou2021manhattan}. The scene understanding field has different branches which focus on different key aspects of the environment. The layout recovery problem has been in the spotlight for many years, obtaining great results with the use of standard and omnidirectional cameras \cite{berenguel2021scaled}\cite{fernandez2020corners}\cite{pintore2020atlantanet}\cite{sun2019horizonnet}. 
This layout information is useful for constricting the movement of autonomous robots \cite{rusli2020roomslam}\cite{salas2015layout} or doing virtual and augmented reality systems.
Another line of research focuses on detecting and identifying objects and their classes in the scene. 
There are many methods for conventional cameras \cite{dvornik2017blitznet}\cite{he2017mask}\cite{russakovsky2015imagenet}, which provide great results, however conventional cameras are limited by their narrow field of view. In recent years, works that use panoramas, usually in the equirectangular projection, are increasing \cite{eder2020tangent}\cite{guerrero2020s}, providing a better understanding of the whole environment.
Besides, the combination of semantic and depth information helps to generate richer representations of indoor environments \cite{koppula2011semantic}\cite{ye2018recurrent}. 
In this work, we focus on obtaining, from equirectangular panoramas, two of the main pillars of scene understanding: semantic segmentation and monocular depth estimation.

\begin{figure}[t]
\centering
\subfloat{
\includegraphics[width=0.23\textwidth]{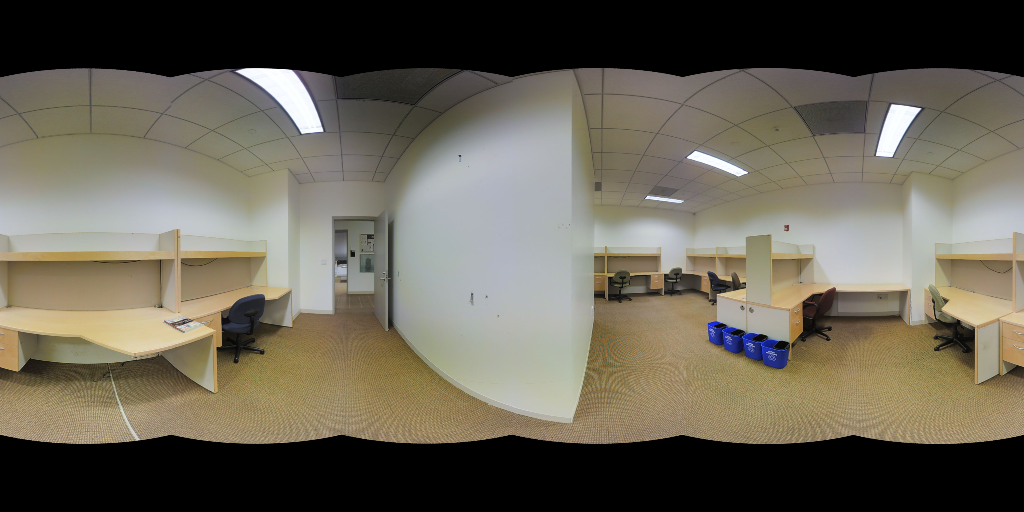}} \hfil
\subfloat{
\includegraphics[width=0.23\textwidth]{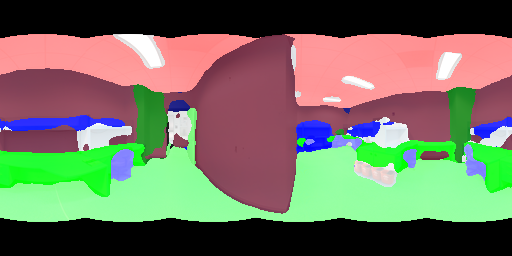}} \\
\subfloat{
\includegraphics[width=0.23\textwidth]{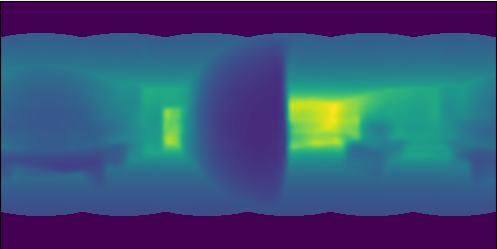}} \hfil
\subfloat{
\includegraphics[width=0.23\textwidth]{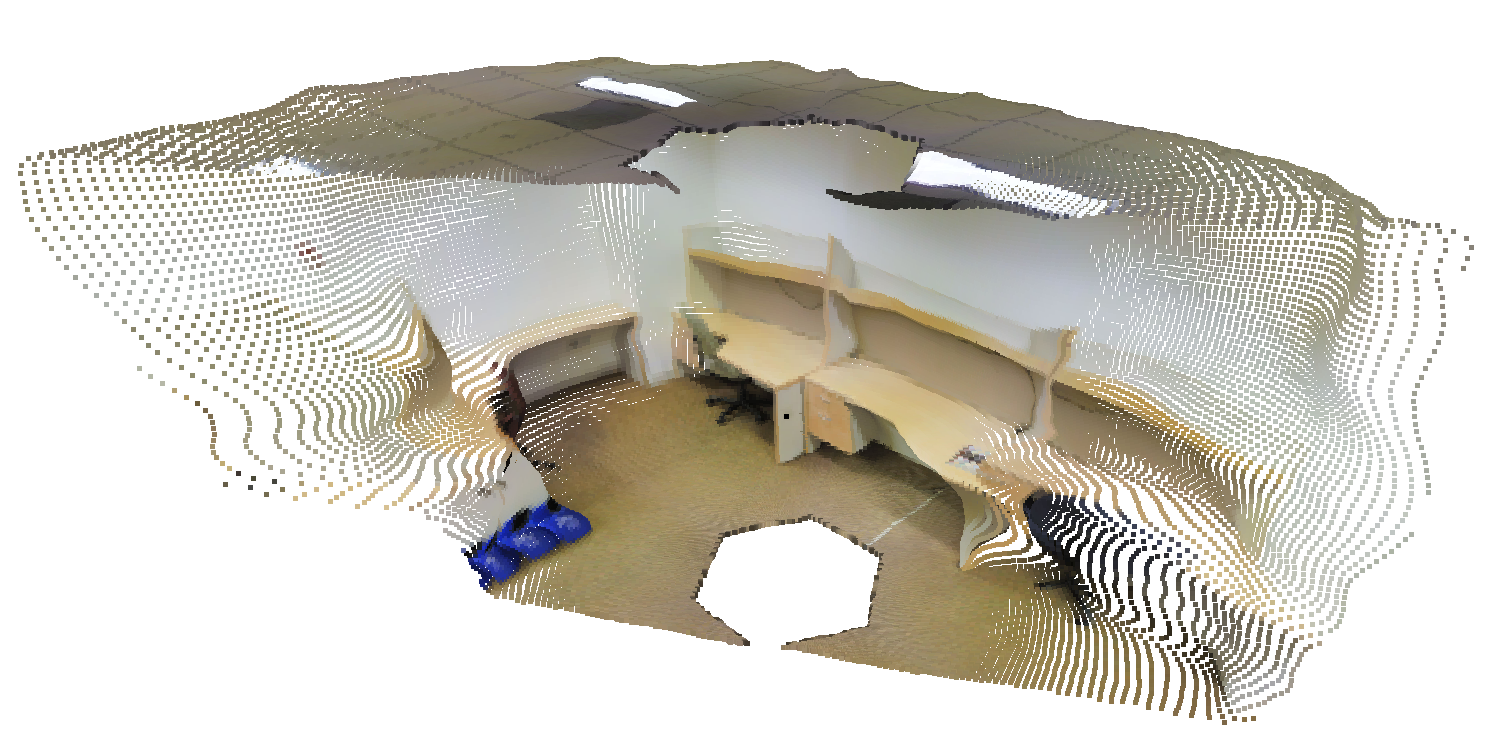}}
\caption{Overview of our proposal. From a single RGB panorama (up left), we make a semantic segmentation (up right) and estimate a depth map (down left) of an indoor environment. With this information we are able to reconstruct in 3D the whole environment (down right).}
\label{fig:intro}
\end{figure}

\begin{figure*}[t]
\centering
\includegraphics[width=1\textwidth]{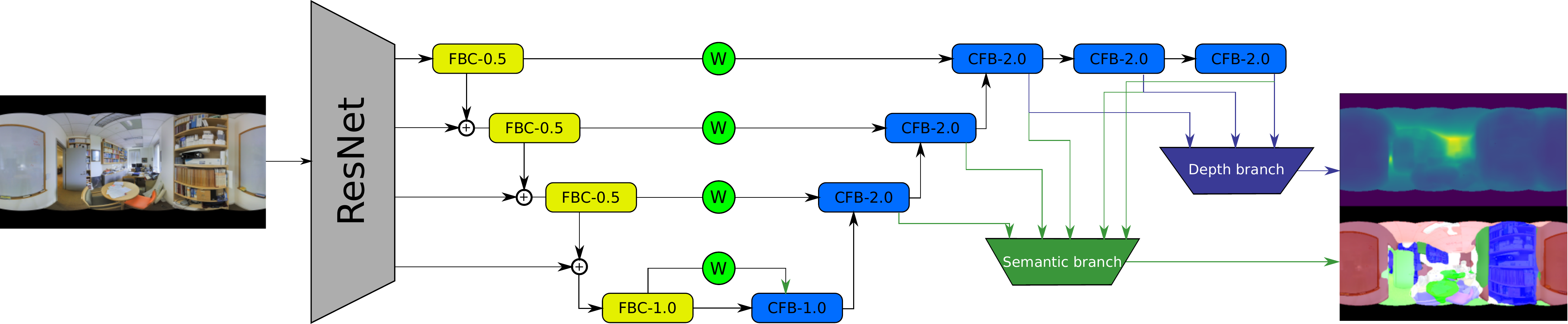}
\caption{Architecture of our {\bf Fre}quential {\bf D}epth estimation and {\bf S}emantic segmentation {\bf Net}work (FreDSNet). The encoder part is formed by a feature extractor (ResNet) and four encoder blocks. The decoder part is formed by six decoding blocks and two branches that predict depth and semantic segmentation. The skip connections from the encoder to the decoder use learned weights.}
\label{fig:modelArch}
\end{figure*}

Without the adequate sensor, navigating autonomous vehicles in unknown environments is an extremely challenging task.
Nowadays there is a great variety of sensors that provide accurate and diverse information of the environment (LIDARs, cameras, microphones, etc.). Among these possibilities, we choose to explore omnidirectional cameras, which have become increasingly popular as main sensor for navigation and interactive applications. These cameras provide RGB information of all the surrounding and, with the use of computer vision or deep learning algorithms, provide rich and useful information of an environment. 

In this paper, we introduce FreDSNet, a new deep neural network which jointly provides semantic segmentation and depth estimation from a single equirectangular panorama (see Fig. \ref{fig:intro}). We propose the use of the fast Fourier convolution (FFC) \cite{chi2020fast} to leverage the wider receptive field of these convolutions and take advantage of the wide field of view of 360 panoramas. Besides, we use a joint training of semantic segmentation and depth, where each task can benefit from the other. Semantic segmentation provides information about the distribution of the objects as well as their boundaries, where usually are hard discontinuities in depth. On the other hand, the depth estimation provides the scene's scale and the location of the objects inside the environment. With this information, we provide accurate enough information for applications as navigation of autonomous vehicles, virtual and augmented reality and scene reconstruction.

The main contribution of this paper is that FreDSNet is the first to jointly obtain semantic segmentation and monocular depth estimation from single panoramas exploiting the FFC. 
The main novelties of our work are: We include and exploit the FCC in a new network architecture for visual scene understanding. Also, we present a fully convolutional neural network that jointly obtains semantic segmentation and depth estimation from single panoramas.


\section{Related works}
\label{sec:related}

\textbf{Semantic segmentation} The semantic segmentation on perspective images is a well-studied field. We can find many works on object detection \cite{russakovsky2015imagenet}, semantic segmentation \cite{he2017mask}\cite{zheng2014dense} or both tasks \cite{dvornik2017blitznet}\cite{girshick2014rich} from perspective cameras. However, omnidirectional images pose a harder problem which is more difficult to tackle. Then, only a few works are able to make object detection or semantic segmentation from omnidirectional images \cite{eder2020tangent}\cite{guerrero2020s}\cite{sun2021hohonet}. 
Since omnidirectional images present heavy distortions (e.g. in spherical projections, like equirectangular images, this distortion is more accentuated in the mapping of the poles) these kinds of images are difficult to manually annotate. Nevertheless, due to the wide field of view of these images (e.g. in the spherical projection, we can see all the surroundings in a single image), the use of omnidirectional images in semantic segmentation is an active field of study since we can obtain a complete semantic understanding of the environment from a single image.

\textbf{Depth estimation} Monocular depth estimation is a research topic that has been on the spotlight in recent years. 
With the rise of deep learning methods, many works on depth estimation from conventional cameras have appeared for diverse applications \cite{facil2019cam}\cite{heo2018monocular}\cite{lee2019monocular}\cite{ranftl2016dense}. Almost at the same time, different works on depth estimation from panoramic images started to appear for indoor scene understanding purposes \cite{pintore2021slicenet}\cite{sun2021hohonet}\cite{wang2020bifuse}\cite{zioulis2018omnidepth}. Each work presents particular approaches for monocular depth estimation, being an open field of study with great interest and many applications. 
 
\textbf{Network architecture} Many recent works on semantic segmentation or depth estimation rely on convolutional encoder-decoder architectures with some recurrent \cite{pintore2021slicenet} or attention mechanism \cite{sun2021hohonet} as hidden representation of the environment. This kind of architectures aim to reduce the spatial resolution of the input image, increasing the number of feature maps in the encoder part, relating the general context of the environment in the hidden representation and up-sampling it in the decoder part to obtain the desired information. 
However, the traditional encoder-decoder architecture which relies on standard convolutions \cite{zioulis2018omnidepth} or geometrical approximations \cite{eder2020tangent} suffers from slow growth of the effective receptive field of the convolutions, losing the general context information that omnidirectional images provide. 

In this work, we propose an encoder-decoder architecture for our network. However, we propose to use the fast Fourier convolution presented in \cite{chi2020fast}, which we denominate Fourier Block since we modify the behaviour of the block. These convolutions have proved that can 'see' the whole image at once, obtaining a higher effective receptive field from early layers. This is a key feature for our proposal since, being aware of the context of the scene, which can only be obtained with omnidirectional images, improves the understanding and interaction with the environment. 

\fancyhf{}
\renewcommand{\headrulewidth}{0pt}
\renewcommand{\footrulewidth}{1pt}
\rfoot{ArXiv Preprint}
\lfoot{\textit{October, 2022}}
\cfoot{\thepage}

\thispagestyle{fancy}

\section{FreDSNet: monocular depth and semantic segmentation}
\label{sec:method}

\begin{figure*}[t]
	\centering
	\includegraphics[width=0.9\textwidth]{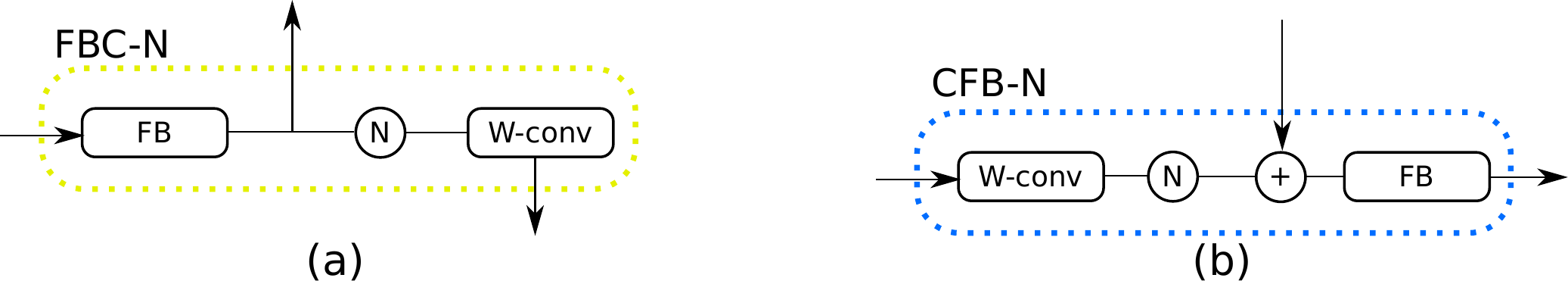}
	\caption{a) FBC-N: encoder block composed by a Fourier Block (FB), skip connection for the Decoder part, Down-Scaling (N) of scale N and a W-Conv (C). b) CFB-N: decoder block composed by a W-Conv (C), Up-Scaling (N) of scale N, addition of a skip connection from the Encoder part and a Fourier Block (FB).}
	\label{fig:modelBlocks}
\end{figure*}

Our network follows an encoder-decoder architecture with Resnet \cite{he2016deep} as initial feature extractor and two separated branches for depth estimation and semantic segmentation. (see Fig.\ref{fig:modelArch}). 
It is inspired by BlitzNet \cite{dvornik2017blitznet} and PanoBlitzNet \cite{guerrero2020s}, using multi-resolution encoding and decoding, in order to obtain a multi-scale representation of the scene, and the use of skip connections, which makes the training process more stable.
Each branch takes intermediate feature maps from the decoder part to provide an output from the multi-scale decoded information. The key novelty of our architecture is how are composed the blocks of encoder and decoder and how these parts are interconnected.

\subsection{Architecture} 

The proposed encoder blocks (FBC-N) are formed by a Fourier Block (FB) followed by a down-scaling (N) and a set of standard convolutions (W-conv) as shown in Fig.\ref{fig:modelBlocks}a. The Fourier Block has the same structure as the FFC implemented in \cite{suvorov2022resolution}, however we differ in the use of the activation function (AF). In the original work, they use a \textit{ReLu} activation function in the FFC (they propose an in-painting method). 
However, recent works as \cite{pintore2021slicenet} have proved that \textit{ReLu} function is not really suited for depth estimation, since it is prone to make gradients vanish. Instead, we use \textit{PReLu} as activation function, which is more stable for monocular depth estimation trainings \cite{pintore2021slicenet}. The same AF change has been made in the \textit{Spectral block}\cite{chi2020fast} from the FB in order to homogenize the behaviour of the network. The output of the FB is the information that we use as skip connection for the decoder part.
After the FB, we rescale the feature map by a factor of N, where N$<$1 means a down-scaling and N$>$1 an up-scaling, followed by a set of standard convolutions defined as W-conv. The W-conv block is defined as 3 consecutive Convolution-Batch normalization-Activation function blocks with circular padding, taking into account the continuity of panoramic images and their features. 
This W-conv block follows the ResNet\cite{he2016deep} \textit{bottleneck} structure for architecture homogeneity. As defined before, we use \textit{PReLu} as activation function. The output of the W-conv is then added, and not concatenated, to the next scale-level output of the feature extractor, as seen in Fig. \ref{fig:modelArch}.

The decoder part follows the same principle as the encoder but in the inverse order, as shown in Fig \ref{fig:modelBlocks}b. The output of the encoder is fed to the decoder blocks (CFB-N) where the feature maps go through a W-conv block and then are up-scaled. The scaled features are added with the corresponding skip connection from the encoder, weighted with a learned parameter, and then go through a FB. The output of each decoder block is then fed to the next decoder block until we recover the full resolution of the network input.

\subsection{Semantic segmentation branch}

From the different feature maps generated by our encoder-decoder architecture, we need to extract the relevant information for the semantic segmentation task. We use the last five sets of feature maps from the decoder. We convolve and upscale each set of features into an intermediate representation. However, instead of concatenating the feature maps, as done in previous works as \cite{dvornik2017blitznet}, we make a learned-weighted sum of them to keep a more compact intermediate representation. Finally, we convolve this intermediate representation into the final representation of the semantic segmentation map. In this branch we switch to a \textit{ReLu} activation function instead of the \textit{PReLu} used in the main body of the architecture. 

\subsection{Depth estimation branch}

For the depth estimation, we have created another separated branch that takes the last three blocks of feature maps from the decoder part. We propose to use these three sets of feature maps since they are the ones with higher resolution, then with higher level of details. As done in the semantic segmentation, we convolve and upscale the features to an intermediate representation, where we add them as a learned-weighted sum. Finally, we convolve again the intermediate representation to make the depth estimation. In this branch, we keep the \textit{PReLu} activation function in the convolution from the feature maps to the intermediate representation. However, we switch to a \textit{ReLu} function in the last convolution since depth cannot be a negative value. We tried different output functions, such as the \textit{PReLu} activation function or not using any, but the \textit{ReLu} function provided the best performance.

\subsection{Loss functions}

Semantic segmentation and depth estimation provide really different information of the environment. However these tasks have common characteristics that can benefit from each other, as the objects boundaries or the room layout \cite{zhang2018joint}. We want to take advantage of these similarities making a joint training where the semantic segmentation and the depth estimation can be jointly predicted.
For our training, we propose to train both branches, segmentation and depth estimation, at the same time and from the same input image. 
For the semantic segmentation loss $L_{Seg}$ , we use the standard \textit{Cross Entropy Loss} and weights for the classes \cite{tian2022striking}, as a solution for the class imbalance in the dataset.

Similar to other state-of-the-art methods for monocular depth estimation \cite{pintore2021slicenet}\cite{wang2020bifuse}, we use an \textit{Adaptive Reverse Huber Loss} (eq. \ref{eq:HuberLoss}) as depth loss function $L_{Dep}$, defined as:
\begin{equation}\label{eq:HuberLoss}
B_c(e) \left\{\begin{matrix}
\left | e \right | & \left | e \right | \leq  c\\ 
\frac{e^2 + c^2}{2c} & \left | e \right | >   c
\end{matrix}\right. ,
\end{equation}\noindent
where $e = Prediction - GroundTruth$ and $c$ is defined as the $20\%$ of the maximum absolute error for each training batch. Following the same idea as \cite{pintore2021slicenet}, we also define the loss function as the sum of the \textit{Adaptive Reverse Huber Loss} on the depth map as well as the gradients (approximated as Sobel Filters). The final $L_{Dep}$ is computed as:
\begin{equation}\label{eq:depLoss}
L_{Dep} = B_{c_1}(\textit{e}) + B_{c_2}(\nabla_x) + B_{c_2}(\nabla_y) ,
\end{equation}\noindent
where $e$ defines the absolute depth error between the prediction and ground truth, $\nabla_x, \nabla_y$ define absolute error between the x, y gradients of the prediction and ground truth respectively, $c_1$ is the threshold in eq. \ref{eq:HuberLoss} for the absolute depth map and $c_2$ is the threshold in eq. \ref{eq:HuberLoss} for the gradients.

In addition to the semantic segmentation and depth estimation standard losses, we add another two losses to help in the joint training process. First, since the range of the depth estimation output should be greater than the semantic segmentation (i.e. the later is closer to a probability distribution while the former is a distance between $0$ and, ideally, infinity), we add a term to force the depth estimation branch to fill the depth range between the closest and farthest pixels. To do so, we compute the mean square difference between the minimum and maximum values of prediction and ground truth in each batch as:
\begin{equation}\label{eq:marginLoss}
L_{mar} = \frac{
\left(y_{gt}^{max} - y_{pred}^{max}\right)^2 + 
\left(y_{gt}^{min} - y_{pred}^{min}\right)^2}
{2} ,
\end{equation}\noindent
where $y_{gt}^{max}$, $y_{pred}^{max}$, $y_{gt}^{min}$ and $y_{pred}^{min}$ are the maximum and minimum values of the ground truth and predicted depth maps respectively. 
Finally, to help in the object boundary definition, we propose to use an object oriented loss $L_{obj}$. This loss helps the network to better define the objects boundaries as well as create the depth discontinuities that appear in these boundaries. To compute the loss, we first compute per-class depth maps from the network prediction and ground truth. Then we compute the mean of the \textit{L1 Loss} of each class depth map to obtain the final $L_{obj}$ as:
\begin{equation}
L_{obj} = \frac{1}{C} \sum_{i=0}^{C} \textit{L1}({y_{gt}}^i,{y_{pred}}^i) ,
\end{equation}\noindent
where $C$ is the number of classes and ${y_{gt}}^i,{y_{pred}}^i$ are the ground truth and predicted class depth maps for the class $i$ respectively.

Our final training loss is the combination of the previous losses. This joint loss function is computed as:
\begin{equation} \label{eq:totalLoss}
L_{total} = \alpha_1 \cdot L_{Seg} + \alpha_2 \cdot L_{Dep} + \alpha_3 \cdot L_{mar} + \alpha_4 \cdot L_{obj} ,
\end{equation} \noindent
where $\alpha_i$ are regularizers to weight the relevance of each individual loss in the final joint loss function. After several tries, we set these regularizers as $\alpha = [9.0,14.0,0.01,5.0]$, obtaining the best performance for our network.

\section{Experiments}
\label{sec:experiments}

In this section we present a set of experiments to validate the joint training method used. We also make a comparison with a state of the art architecture for depth estimation and semantic segmentation. Finally, we present qualitative results and application examples for our network.

To evaluate our work and compare it with the state of the art, we use the following metrics. For the depth estimation task, we use the standard metrics introduced by \cite{zioulis2018omnidepth}. We use the Mean Relative Error (\textit{MRE}), Mean Absolute Error (\textit{MAE}), Root-Mean Square Error of linear (\textit{RMSE}) and logarithmic (\textit{RMSElog}) measures, and three relative accuracy measures defined as the fraction of pixels where the relative error is within a threshold of $1.25^n$ for $n=1,2,3$ (\textit{$\delta^1,\delta^2,\delta^3$}). On the other hand, for the semantic segmentation task, we use standard metrics as the mean Intersection over Union (\textit{mIoU}), computed as the average \textit{IoU} for each class except the \textit{Unknown} class; and the mean Accuracy (\textit{mAcc}), computed as the average accuracy for each class except the \textit{Unknown} class.

\subsection{Ablation study}

\begin{table}[t]
\caption{Ablation study of Loss functions. Best validation metrics obtained on each training.}
\centering
	\begin{tabular}{cccc|cc|cc}
	$L_{Seg}$	&	$L_{Dep}$	&	$L_{mar}$	&	$L_{obj}$	 & 	
	MRE & MAE & 
	mIoU  & mAcc \\ \hline \hline
	\checkmark	& \checkmark	& $\times$		& $\times$		& 
	0.0613 		& 0.0950		& 60.3			& 81.9	\\ \hline 
	\checkmark	& \checkmark	& $\times$		& \checkmark	& 
	0.0583 		& 0.0855		& 61.5			& 82.5	\\ \hline 
	\checkmark	& \checkmark	&  \checkmark	& $\times$		&
	0.0560 		& 0.0898		& 60.6			& 81.6	\\ \hline 
	\checkmark	& \checkmark	& \checkmark	& \checkmark	& 
	\textbf{0.0553}& \textbf{0.0827}& \textbf{62.7}	& \textbf{84.2}	\\ \hline
	\multicolumn{4}{r}{}
	& \multicolumn{2}{c}{\scriptsize{\textit{Lower is better}}}
	& \multicolumn{2}{c}{\scriptsize{\textit{Higher is better}}}
	
	\end{tabular}
\label{tab:lossAblation}
\end{table}

\begin{table}[t]
\caption{Ablation study of joint training. Best validation weights are used in each of the cases for the evaluation.}
\centering
	\begin{tabular}{l|cc|cc}
	Training		 & 	
	MRE & MAE  & 
	mIoU  & mAcc 	\\ \hline \hline
	Depth	& 0.1401 & 0.1773& - 	& - 	\\\hline
	Semantic& - 	& - 	 & 40.3	& 55.1	\\ \hline 
	Joint	& \textbf{0.0952}& \textbf{0.1227}& \textbf{46.1}	& \textbf{63.1} \\ \hline
	\multicolumn{1}{r}{}
	& \multicolumn{2}{c}{\scriptsize{\textit{Lower is better}}}
	& \multicolumn{2}{c}{\scriptsize{\textit{Higher is better}}}
	\end{tabular}
\label{tab:trainAblation}
\end{table}

\begin{table*}[t]
\centering
\caption{Quantitative comparison for Depth Estimation and Semantic Segmentation on the Stanford 2D3DS dataset \cite{armeni2017joint}.}
\begin{tabular*}{0.9\textwidth}{@{\extracolsep{\fill}} c||ccccccc||cc}

	Network &
	MRE $\downarrow$ 	& MAE $\downarrow$ 	  &
	RMSE $\downarrow$ 	& RMSElog $\downarrow$  &
	$\delta^1 \uparrow$ & $\delta^2 \uparrow$ & $\delta^3 \uparrow$ &
	mIoU $\uparrow$ & mAcc $\uparrow$\\ \hline
	HoHoNet\cite{sun2021hohonet} &
	0.1124 & 0.2265 & 0.4027 & 0.0710 & 0.8994 & 0.9687 & 0.9879 & 30.7 & 40.5 \\
	{\bf Ours} &
	0.0952 & 0.1327 & 0.2727 & 0.0436 & 0.8424 & 0.9583 & 0.9863 & 46.1 & 63.1 \\
	\hline
\end{tabular*}
\label{tab:compExp}
\end{table*}
In this first subsection, we substantiate the joint training of depth and semantics and also evaluate the use of a combined loss function. 

\textbf{Loss Function}. 
We evaluate how the overall performance of our network is affected by each loss function. For that purpose, we perform the same joint training using the task-specific losses and adding sequentially the new losses that we propose in Section \ref{sec:method}-D. We evaluate the performance of the network with four different metrics: two for depth estimation, \textit{MRE} and \textit{MAE}; and two for semantic segmentation, \textit{mIoU} and \textit{mAcc}.
The results from Table \ref{tab:lossAblation} show that the use of these new losses increase the performance of our network. 

\textbf{Joint training}. On a second step, we show that the joint training of depth estimation and semantic segmentation benefits the overall performance of the network.
For this purpose, we train our network for task-specific purposes, that means with only one of the branches at a time, and in the proposed joint training, with both branches. Notice that only half of the metrics are used for the task-specific trainings, since the other half correspond to a different task. In the task-specific trainings, we only use the specific loss function for each task, i.e. we only use $L_{Dep}$ for the specific depth estimation training and $L_{Seg}$ for the specific semantic segmentation training. We use the same metrics as in the previous experiment to compare the performance of the different trainings. 
The results presented on Table \ref{tab:trainAblation} confirm our first intuition, which confirm the results of \cite{zhang2018joint}. The joint training of semantic segmentation and depth estimation increases the performance of the network for both tasks.

\subsection{State of the art comparison}
We compare our work with the sole state of the art method which obtains semantic segmentation and depth estimation HohoNet \cite{sun2021hohonet}. We train this network in the same conditions as our network and compare the performance of both proposals. 
We use the Stanford2D3DS Dataset \cite{armeni2017joint} with the first folder split, which uses \textit{Area 5} as test set and the other areas for training and validation. We can only use the Stanford2D3DS dataset for both tasks since it is the only public dataset with semantic and depth information for real equirectangular panoramas.

\begin{figure*}[h!]
\centering
\subfloat{\includegraphics[width=0.24\textwidth]{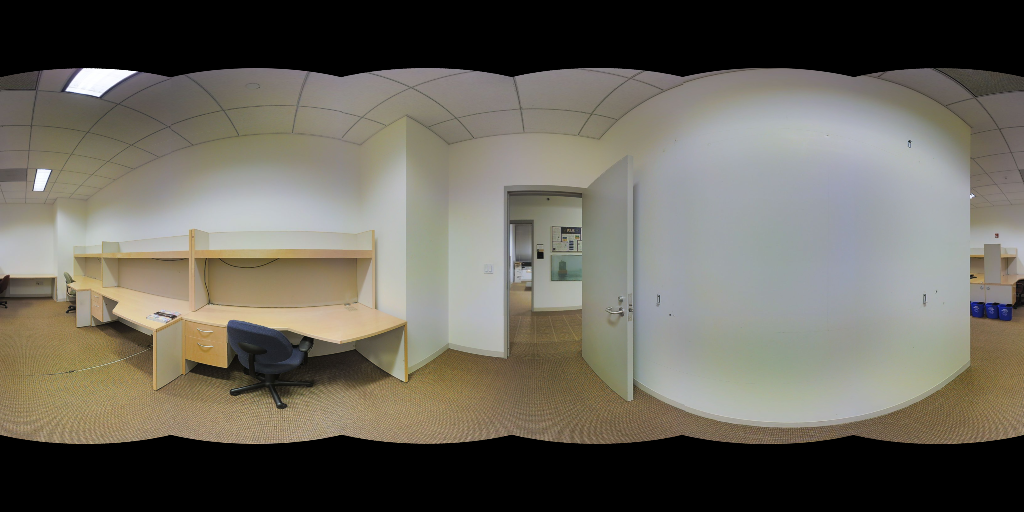}}\hfil
\subfloat{\includegraphics[width=0.24\textwidth]{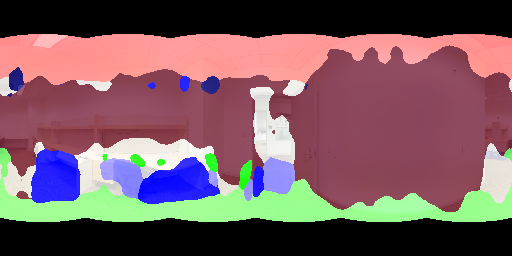}}\hfil
\subfloat{\includegraphics[width=0.24\textwidth]{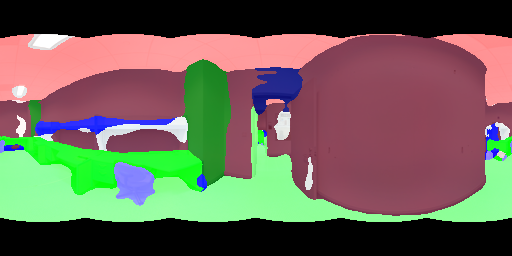}}\hfil
\subfloat{\includegraphics[width=0.24\textwidth]{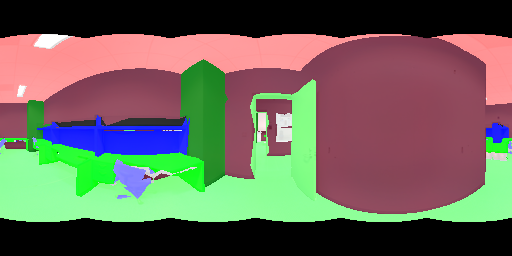}}
\\ \vspace{-2mm}
\subfloat{\includegraphics[width=0.24\textwidth]{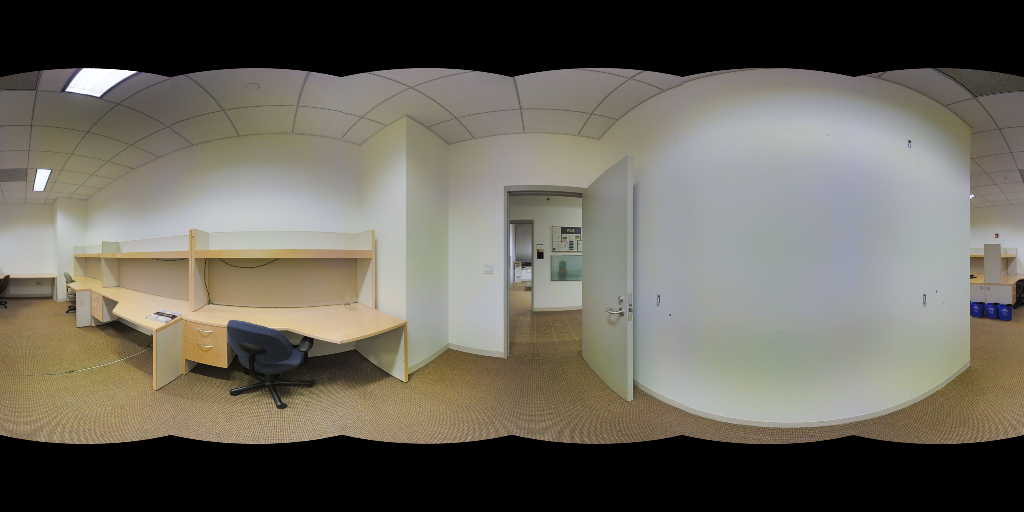}}\hfil
\subfloat{\includegraphics[width=0.24\textwidth]{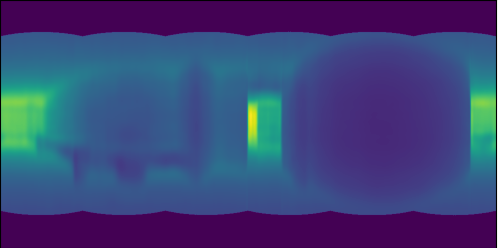}}\hfil
\subfloat{\includegraphics[width=0.24\textwidth]{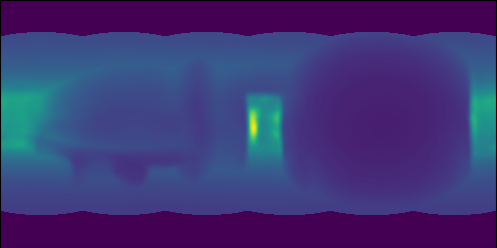}}\hfil
\subfloat{\includegraphics[width=0.24\textwidth]{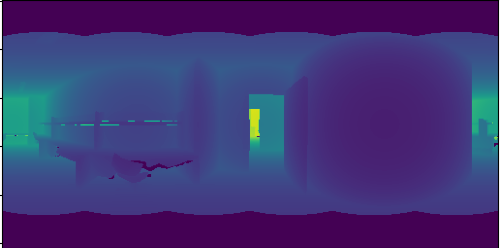}}
\\ \vspace{-2mm}
\subfloat{\includegraphics[width=0.24\textwidth]{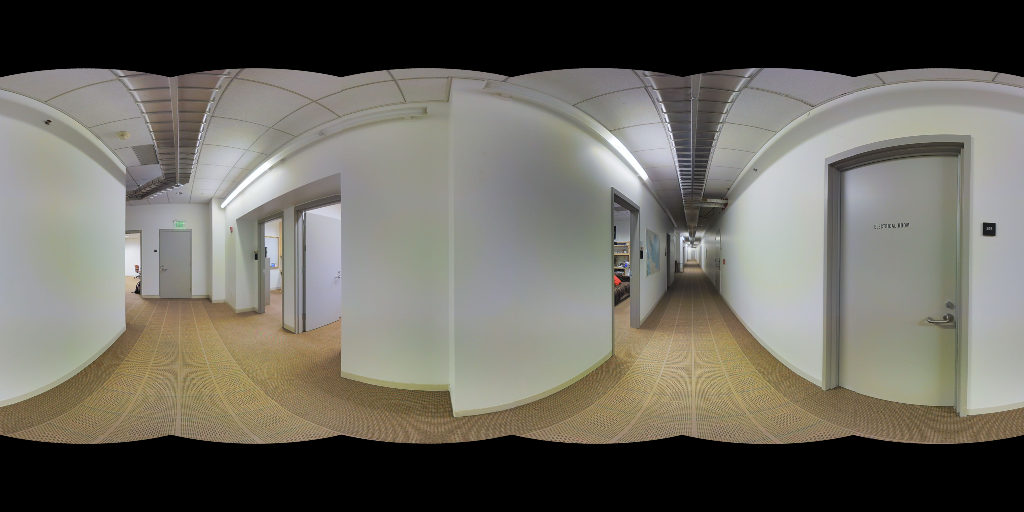}}\hfil
\subfloat{\includegraphics[width=0.24\textwidth]{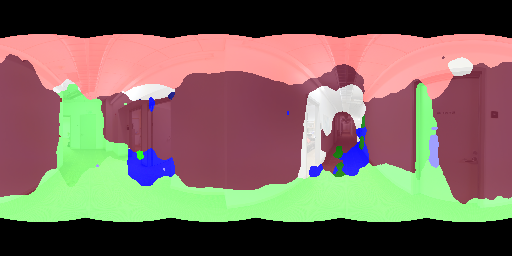}}\hfil
\subfloat{\includegraphics[width=0.24\textwidth]{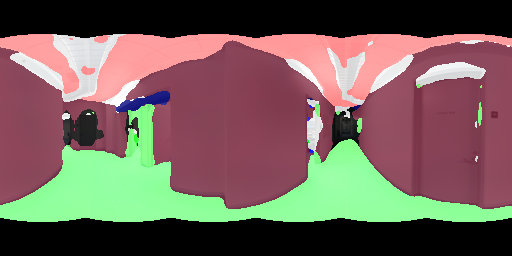}}\hfil
\subfloat{\includegraphics[width=0.24\textwidth]{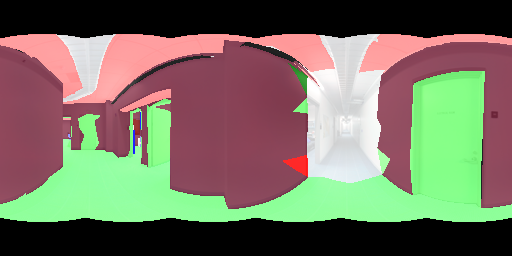}}
\\ \vspace{-2mm}
\subfloat[Input]
{\includegraphics[width=0.24\textwidth]{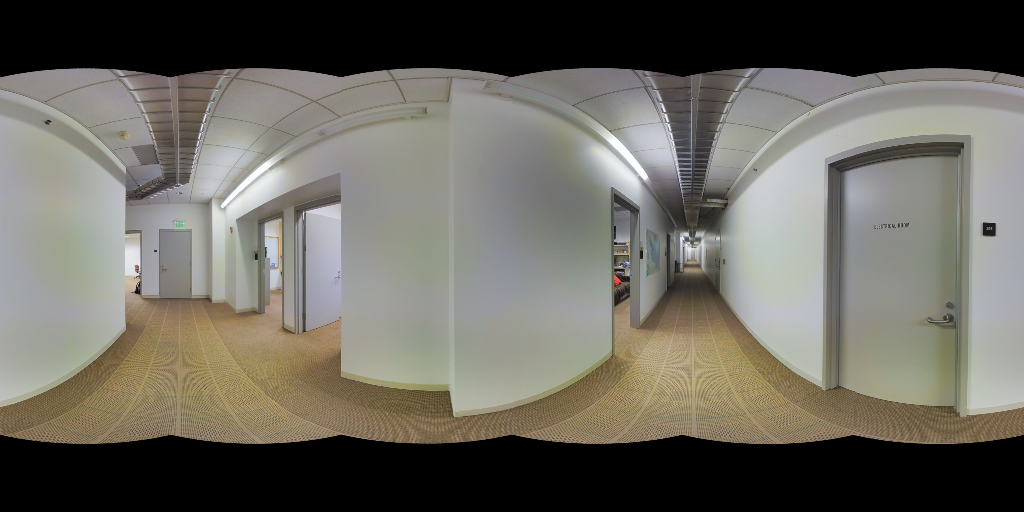}}\hfil
\subfloat[HohoNet]
{\includegraphics[width=0.24\textwidth]{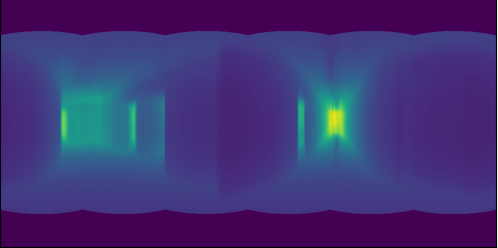}}\hfil
\subfloat[\textbf{Ours}]
{\includegraphics[width=0.24\textwidth]{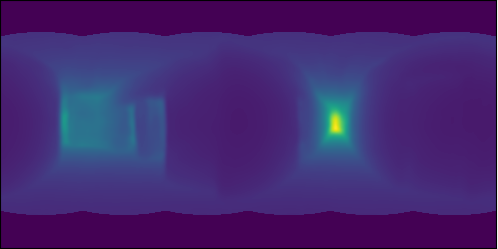}}\hfil
\subfloat[Ground Truth]
{\includegraphics[width=0.24\textwidth]{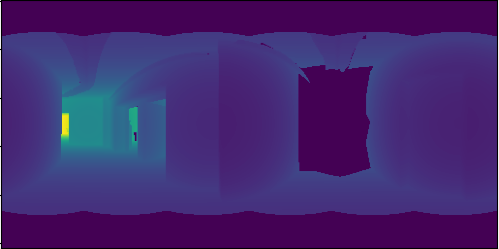}}\hfil
\caption{Qualitative comparison between HohoNet \cite{sun2021hohonet} and our proposal for semantic segmentation and depth estimation in Stanford2D3DS \cite{armeni2017joint}.}
\label{fig:Comparison}
\end{figure*}

We train both networks in one GPU Nvidia RTX2080-Ti. The initial learning rate of our training is set to $1e-5$ and we use exponential decay and periodic step reductions in the learning rate. For HohoNet\cite{sun2021hohonet}, we use their available code for training. The quantitative results of the evaluation are presented in Table \ref{tab:compExp}. We also present qualitative results and comparison of both networks in Figure \ref{fig:Comparison}.

An important remark is that our network provides, at the same time, depth and semantic information while HohoNet presents two slightly different architectures for task-specific solutions. 
That means, HohoNet only provides depth or semantics one by one, needing two different trainings and two evaluations for the two tasks while we only need to train and evaluate our network once for both tasks.

The quantitative comparison shows that our network performs similar to specific state of the art methods under the same conditions. The results from HohoNet presented in this paper differ from the originals \cite{sun2021hohonet} since the training conditions are different. In this case, we only train both networks in the Stanford Dataset, without pre-trained weights on other datasets for the whole network, using the \textit{Area 5} only at test and not during the training or validation steps. In these conditions, FreDSNet performs similar in the monocular depth estimation and slightly better in the semantic segmentation.

\captionsetup[subfigure]{labelformat=empty, labelsep=none}

\subsection{Scene understanding for navigation}
In this third subsection we present different results from our network and ideas of applications. With the combination of semantics and depth, we can extract the free space for navigation (extracting the floor) or compute where the obstacles are located (computing the position of the different segmented objects).

Navigation algorithms for mobile robots require to detect the obstacles and the free space around the vehicle. With an omnidirectional camera we can obtain RGB information of the surroundings in one shot. Then, FreDSNet can simultaneously obtain a semantic and depth maps of the environment. With this information, we can obtain different useful representations of the scene, allowing a better interaction of a mobile vehicle with the environment allowing a robot to be autonomous in unknown environments.
Also, in the case of autonomous vehicles, it is important to be able to work in real time. 
We have evaluated the feasibility of our network for such a task obtaining that it can work at 33 frames per second with panoramas of $512\times256$ pixels of resolution (average speed computed with the test set of the dataset).

\begin{figure*}
\centering
\subfloat[RGB]
{\includegraphics[width=0.32\textwidth]{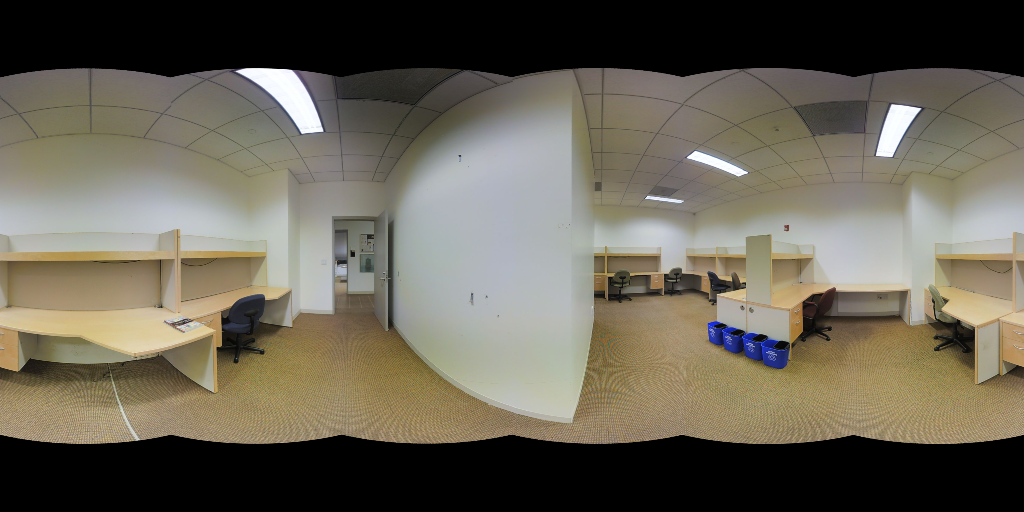}}\hfil
\subfloat[Semantic Segmentation]
{\includegraphics[width=0.32\textwidth]{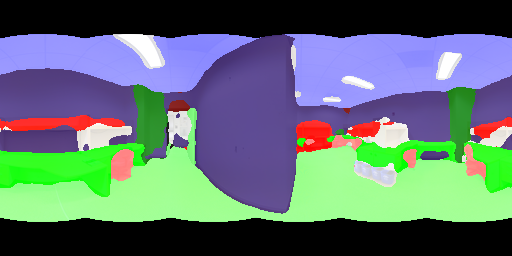}} \hfil
\subfloat[Depth Estimation]
{\includegraphics[width=0.32\textwidth]{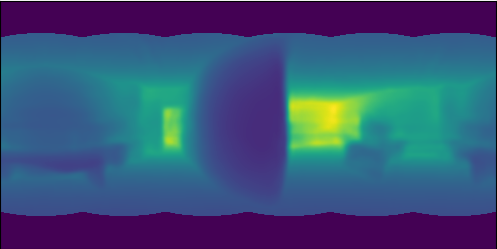}} \\
\subfloat[Free Floor]
{\includegraphics[width=0.24\textwidth]{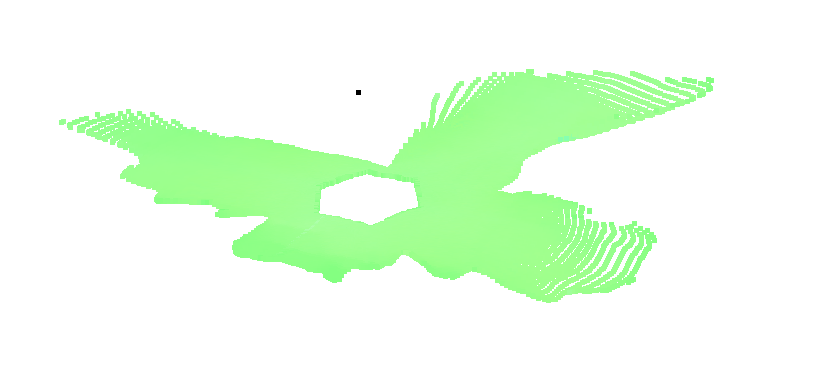}}\hfil
\subfloat[Room structure]
{\includegraphics[width=0.24\textwidth]{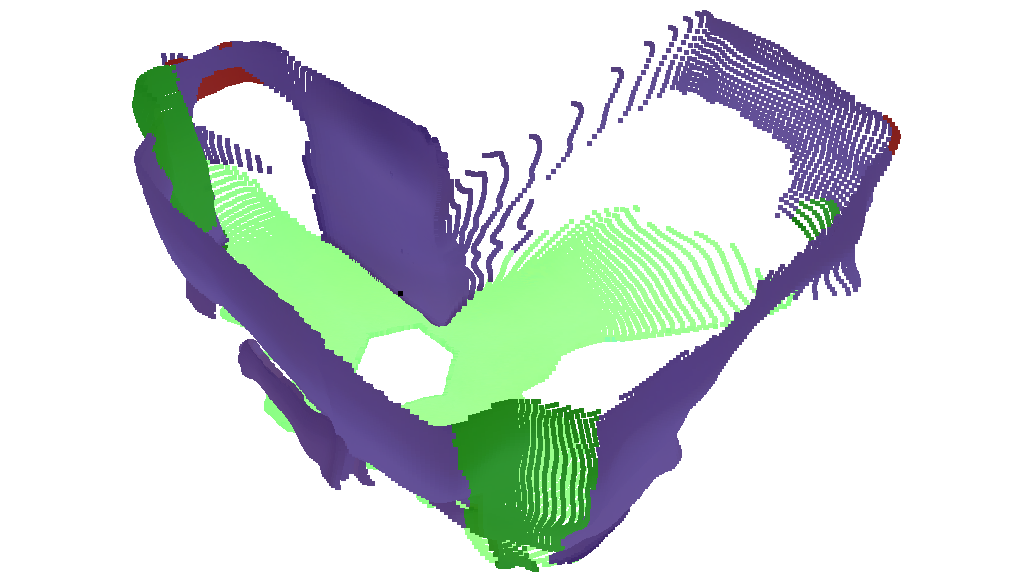}}\hfil
\subfloat[Room and obstacles]
{\includegraphics[width=0.24\textwidth]{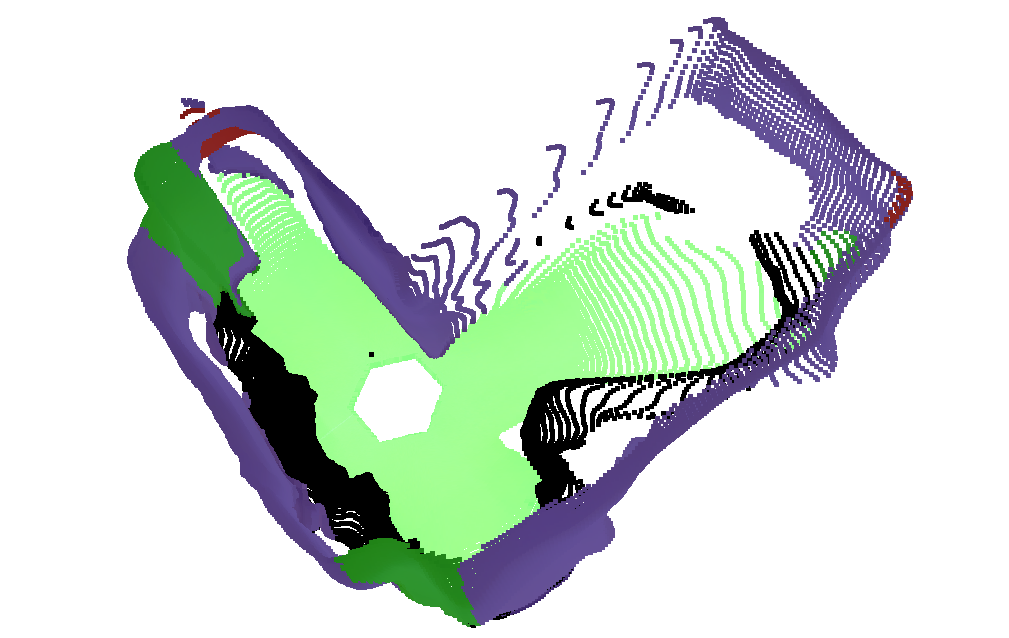}}\hfil
\subfloat[Semantic reconstruction]
{\includegraphics[width=0.24\textwidth]{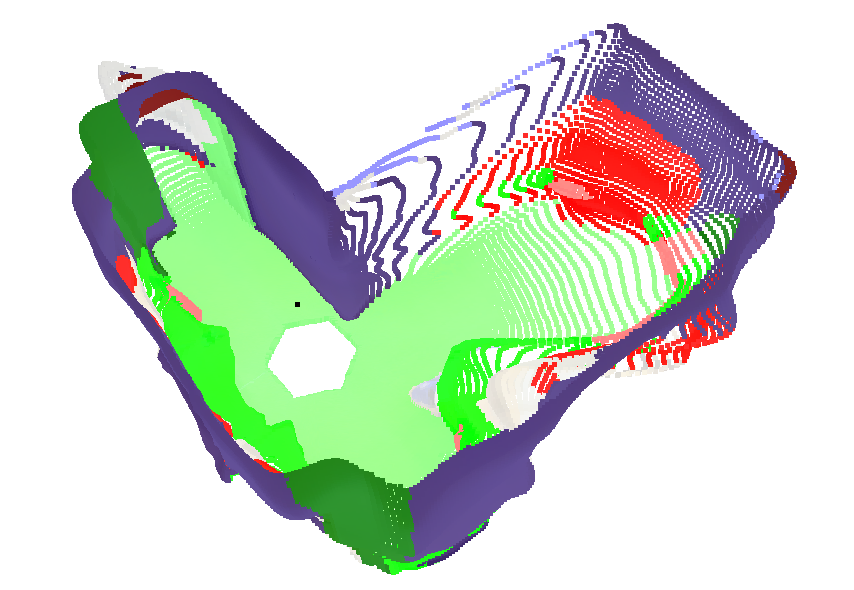}}\hfil
\caption{In the first row: RGB is the input of our network which outputs the Semantic Segmentation and Depth estimation. In the second row: different useful environment representations that can be obtained from the output information provided by FreDSNet. (For a better representation, the ceiling has been removed from all visualizations)}
\label{fig:navigation}
\end{figure*}

As an example of the information that can be obtained from our network, in Figure \ref{fig:navigation} we present several useful environment representations from a single equirectangular panorama. From left to right in the Figure, we show: free floor reconstruction, thought for terrestrial	robot navigation; Room structure reconstruction, defines the maximum space that a vehicle can move; Room and obstacles, includes the room structure and the obstacles (in black) of the room; and the semantic reconstruction, which defines the environment and the different objects with which an autonomous vehicle can interact in the environment.

\section{Conclusion}
\label{sec:conclusion}

We have presented FreDSNet, a neural network for joint monocular depth estimation and semantic segmentation from single equirectangular panoramas. Our network is the first that exploits convolutions in the frequential domain for scene understanding. 
Also, we have proposed a joint training which improves the performance for both tasks, which has been validated experimentally on the Stanford2D3DS dataset. 

The experiments performed validate our proposed contributions: the FFC is a good asset for indoor scene understanding allowing better understanding from more simple network architectures and the joint training mutually benefits the performance of both tasks. 
Besides, the comparison made shows that we provide results in the state of the art for monocular depth estimation and semantic segmentation networks for equirectangular panoramas.

The research of scene understanding methods is still an open topic. Many different approaches are appearing, making improvements in key aspects. In this work we provide a novel solution which information can be used in many others research fields such as virtual or augmented reality, robot navigation and interaction with the environment.

\section*{ACKNOWLEDGEMENT}
This work was supported by projects RTI2018-096903-B-100 (AEI/ FEDER, UE) and JIUZ-2021-TEC-01.


\bibliographystyle{ieeetranS}
\bibliography{mybib}

\end{document}